\documentclass[conference]{IEEEtran}
\pdfoutput=6
\IEEEoverridecommandlockouts
\usepackage{cite}
\usepackage{amsmath,amssymb,amsfonts}
\usepackage{algorithmic}
\usepackage{graphicx}
\usepackage{hyperref}
\usepackage{textcomp}
\usepackage{xcolor}
\usepackage{pdfpages}
\def\BibTeX{{\rm B\kern-.05em{\sc i\kern-.025em b}\kern-.08em
    T\kern-.1667em\lower.7ex\hbox{E}\kern-.125emX}}
\begin{document}

\title{A Machine Learning Approach to Digital Contact Tracing: TC4TL Challenge\\
{\footnotesize}
\thanks{}
}

\author{
\IEEEauthorblockN{ Badrinath Singhal }
\IEEEauthorblockA{\textit{ School of Computer Science } \\
\textit{University College Dublin}\\
Dublin, Ireland \\
 badrinath.singhal@ucdconnect.ie }
\and 
\IEEEauthorblockN{ Chris Vorster } 
\IEEEauthorblockA{\textit{ School of Electronic Engineering } \\
\textit{Dublin City University}\\
 Dublin, Ireland \\
 work@chrisvorster.co.za } 

\and
\IEEEauthorblockN{Di Meng}
\IEEEauthorblockA{\textit{ School of Computer Science } \\
\textit{University College Dublin}\\
Dublin, Ireland \\
di.meng@ucdconnect.ie }

\and  
\IEEEauthorblockN{\qquad \qquad  Gargi Gupta}
\IEEEauthorblockA{\textit{\qquad \qquad School of Computer Science } \\
\textit{\qquad \qquad  TU Dublin}\\
\qquad \qquad Dublin, Ireland \\
\qquad \qquad D21125205@mytudublin.ie }

\and
\IEEEauthorblockN{\qquad \qquad Laura Dunne}
\IEEEauthorblockA{\textit{\qquad \qquad School of Computer Science } \\
\textit{\qquad \qquad University College Dublin}\\
\qquad \qquad Dublin, Ireland \\
\qquad \qquad laura.dunne2@ucdconnect.ie}
\and
\IEEEauthorblockN{\qquad \qquad Mark Germaine}
\IEEEauthorblockA{\textit{\qquad \qquad School of Electronic Engineering } \\
\textit{\qquad \qquad Dublin City University}\\
\qquad \qquad Dublin, Ireland \\
\qquad \qquad mark.germaine2@mail.dcu.ie}
}
\maketitle
\begin{abstract}
Contact tracing is a method used by public health organisations to try prevent the spread of infectious diseases in the community. Traditionally performed by manual contact tracers, more recently the use of apps have been considered utilising phone sensor data to determine the distance between two phones. In this paper, we investigate the development of machine learning approaches to determine the distance between two mobile phone devices using Bluetooth Low Energy, sensory data and meta data. We use TableNet architecture and feature engineering to improve on the existing state of the art (total nDCF 0.21 vs 2.08), significantly outperforming existing models.
\end{abstract}


\section{Introduction}
On $29^{th}$ of December 2019, an outbreak of pneumonia of unknown aetiology was reported in Wuhan City, Hubei, in China to the World Health Organisation (WHO) \cite{b1}. This viral agent was named as Severe Acute Respiratory Syndrome Coronavirus 2 (SARS-CoV-2), with the disease it induced name COVID-19 given by the WHO \cite{b1}. The WHO later declared the outbreak of COVID-19 to be an international public health emergency on the $30^{th}$ of January 2020, and as of today countries around the world continue to deploy various public health measures such as basic hand washing, use of sanitizers and masks, implementing lockdowns, to the use of artificial intelligence-powered chatbots to mitigate the spread of the disease\cite{b3}\cite{b4}. 

To curb the effect of pandemic, people stayed and started working from home which saw a rise in the use of mobile applications during this period. David ME et al. in their study highlighted increase in use of smartphones during pandemic \cite{b15}. Henceforth, in addition to the prominent measures taken by governments all over the world, to disseminate information through various platforms one of the major concerns of scientists across the world is to determine how technology and smartphones can be used for COVID-19 crisis management. 

The WHO identified contact tracing as a necessary tool for pandemic control \cite{b16}. Contact tracing helps us to identify who has been in “close contact” with a person infected with SARS-CoV-2 and the person in contact can be notified by automatic messages via smartphones to get tested or restrict their movement for a period \cite{b5}. The definition of a close contact varies by region, but the European Centre for Disease Prevention and Control defines it as "people who have been within 2 metres of each other for 15 minutes or more” \cite{b8}. The EU Commission also recommended that contact tracing is helpful to track the people who return to hotels and camping sites\cite{b16} and a prerequisite for reopening of various places across different countries. Limitation of human contact tracing motivated various governments to partner with IT industries to deploy digital contact tracing. Various approaches are currently being used by various tech industries, but in this study, we will focus on proximity-based contact tracing using bluetooth data and other mobile sensor data. Our focus is also on determining the role of machine learning in contact tracing.

Along with traditional contact tracing methods, in some countries like South Korea contact tracing involves checking CCTV, GPS location and credit card transactions. In the European Union, such measures breach the GDPR rules and could not be rolled out. However, what has emerged is “opt in” smartphone-based contact tracing apps. These apps rely on the emitted Bluetooth low energy signals (BLE), or “chirps” from the smartphones, which other phones will detect and record the pseudo-random bit sequence emitted by the chirping phone as well as the other parameters such as estimated transmission power of the emitting phone and receiving power of the receiving phone. \cite{b9}\cite{b10}. However, the received signal strength indicator (RSSI) value of Bluetooth chirps sent between phones is a very noisy estimator of the actual distance between the phones and can be dramatically affected in real-world conditions by i) where the phones are carried, ii) body positions, ii) physical barriers, and iv) multi-path environments, and many others \cite{b19}. Leith et al. in their study found very little correlation between RSSI and distance between smartphones while on the tram in Dublin, potentially due to the reflections on the metal structure of the trams which can act as a reflector of radio signals\cite{b11} \cite{b12}. Leith et al. in their research also observed that the BLE RSSI can vary dramatically depending on the relative orientation of the handsets and on absorption by the body\cite{b13}. It was also identified that people sitting with their phone in their pocket received low signal strength despite sitting within 1m. This appears to be less of an issue in outdoor settings where RSSI tends to attenuate with increasing distance \cite{b13}. Hatke et al.  proposed that acknowledging the phone carriage state of both the transmitting and receiving devices would improve the reliability of using BLE, such as knowing whether the phone was in the pocket or in the hand\cite{b14}.

National Institute of Standards and Technology (NIST) proposed a challenge to people to devise a machine learning solution to identifying people who were “Too close for too long” (TC4TL), tasking people to come up with solutions using machine learning and RSSI values combined with other features supplied in datasets provided provided by them. Further, challenge is described in detail in Section II. 
The primary objective of this study is to accurately predict the distance between two mobile phone devices using mobile phone sensor data and Bluetooth Low energy whilst improving the current state of the art.

This paper is structured as follows in Section II we discuss the challenge description and the dataset used for this experiment. Section III discusses background study for this research study. Section IV describes the methodology used to conduct this experiment. Results and discussions about the study conducted is discussed in Section V. With few closing remarks and future scope sections VI and VII , concludes the paper. 

\section{Challenge Description \& Dataset Used}
\subsection{Challenge Description}
Too Close for Too Long (TC4TL) challenge was announced by NIST in collaboration with MIT Private Automated Contact Tracing (PACT) research group \cite{b10}\cite{b19}. The primary objective of this challenge is to estimate the distance and time between the two phones given a series of RSSI values along with the other sensor data. The challenge given by SFI Centre for Machine Learning (ML-Labs) primarily focuses on to identify the role of machine learning in the estimation of distance between the two phones, while estimation of time is not the part of challenge. The main aim of this challenge is 1) to explore promising new ideas and identify role of machine learning in TC4TL detection using BLE signals. 2) support the development of advanced technologies incorporating these ideas and 3) to measure performance of the state-of-the-art TC4TL detectors \cite{b19}. It should be noted, the models in this paper were optimised to fulfill the purpose of this challenge as evaluated by the software provided by NIST.
\subsection{Dataset Description and Dataset Used }
The data provided by NIST for this challenge consists of training datasets which consists of $15,552$ event files, development datasets (936 event files) and test datasets which contains $8,423$ event files. Additionally, the PACT GitHub repository also provides access to additional datasets that could be used to train models. According to the TC4TL Evaluation Plan \cite{b22} the \emph{development dataset} and any other dataset except the \emph{MIT Matrix Dataset} can be used for training. For this research study the NIST \emph{development dataset} and the MITRE Range Angle (Structured) dataset are used to train the model. To understand the challenge better it is important to know how the data was collected.

The data was collected with one tester (beacon) staying stationary at one end of the test space at position 0 (figure 1). Another tester (receiver) runs a contact tracing app on their phone to receive the signal from the beacon. The receiver moved along a straight-line distance from the beacon, monotonically, at marked distances as illustrated in Figure 1. The corresponding distance in metres is $0.9, 1.2, 1.5, 1.8, 2.4, 2.7, 3.0, 3.6$ and $4.5 m$. For the purposes of this challenge we consider $0.9-1.2 = 1.2 m, 1.5-1.8 = 1.8 m, 2.4-3.0 = 3 m$ and $3.6-4.5 = 4.5 m$. Starting at $3$ ft, the receiver starts the test facing the beacon and the phone location of the devices is saved. Every $15$ seconds, the receiver rotates their position $45$ degrees until all $8$ angles have been collected. The receiver then repeats this process for the remaining distances and the test is complete.

\begin{figure}[htbp]
\centerline{\includegraphics[scale=0.15]{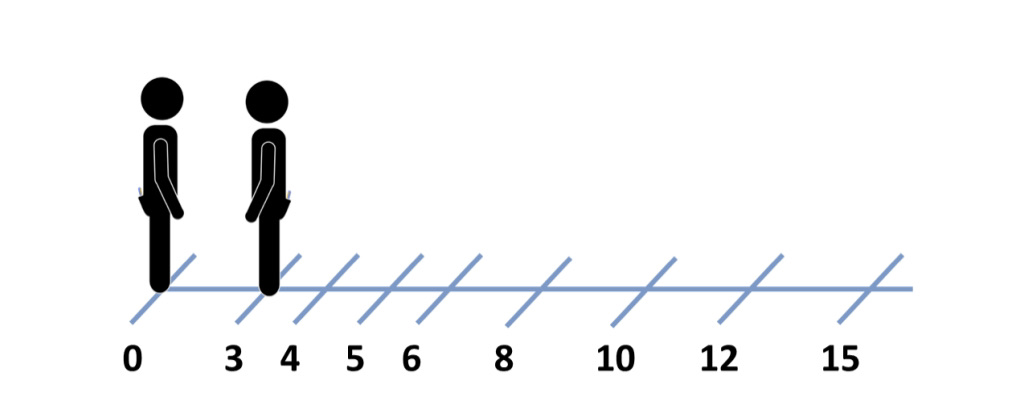}}
\caption{Distances at which data was collected in feet}
\label{fig}
\end{figure}
In the dataset within each event file, the receiver records the data in 4 second windows called “looks”, with varying step sizes in between each look. Therefore we speculate the last look(s) in each event file should correlate with the furthest distance from the beacon in a given event, as we know the data is collected monotonically.
Further the data can be subset into coarse grained ($1.8 m$ or $4.5 m$) or fine grain events ($1.2, 1.8, 3.0$ \& $4.5 m$). As such, coarse grain events had two possible targets to predict whereas fine grain events had four possible targets to predict, thus this was treated as a classification problem. Both the coarse grain and fine grain data were balanced in terms of target outcome.
\subsection{Challenge Evaluation }
The challenge submissions are evaluated locally using the evaluation software provided by NIST. The distance and time estimates are converted into contact event hypothesis labels, i.e., TC4TL or not-TC4TL, using two parameters distance ($D$) and time ($T$). Since we will not be estimating time component of TC4TL detector $T$ will be set as $T=0$. To measured the model's performance parameters such as probability of false negative ($P_{miss}$) and probability of false positive ($P_{fa}$) are calculated. A normalized decision cost function ($nDCF$) combines these two errors into a single value using weights reflecting the relative cost of each type of error. 
The reference and hypothesized event types are compared over a set of contact events and the probability of miss ($P_{miss}$) and probability of false alarm ($P_{fa}$) values are calculated. Mathematically above statement can be summarized as: 

\begin{equation}
    P_{miss} = \frac{Num of ref=TC4TL and hyp=not - TC4TL}{ref = TC4TL}
\end{equation}
\begin{equation}
    P_{fa} = \frac{Num of ref=not - TC4TL and hyp=TC4TL}{ref=not-TC4TL}
\end{equation}
\begin{equation}
    nDCF = \frac{w_{miss}P_{miss} + w_{fa}P_{fa}}{min(w_{miss}, w_{fa})}
\end{equation}

\section{Related Work }
Abbar et al.  in their paper discussed the various digital approaches useful in contact tracing and highlighted the impact and role of AI in contact tracing\cite{b16}.

Gómez C et al. in their research experiment developed two machine learning models GBM and MLP to estimate the distance between the two phones on the dataset provided by NIST for TC4TL challenge. The problem was considered as classification task with coarse grain and fine grain as the labels. It was also identified that feature selection and a handcrafted feature called as estimated distance effects the accuracy scores of the model\cite{b17}. Their approach outperformed the top model presented at the NIST challenge by Hong Kong University of Science \& Technology (HKUST) who used Bluetooth signal, Inertial Measurement Unit (IMU), and phone transmission power (TxPower) information and used Deep Neural Network (DNN) as the Classification model\cite{b20}. It was observed that proposed Gradient Boosting Machine (GBM) and Multi-layer Perceptron (MLP) models gave better average nDCF value of $0.5175$ and $0.52$ respectively as compared to average value of $nDCF$ $0.555$ produced by the HKUST. 

Shankar et al. in their research study approached the NIST TC4TL challenge to predict the distance between the two phones using the phones sensor data i.e., received signal strength indicator (RSSI) from the bluetooth low energy (BLE) and other features that impact the RSSI values. The models were trained on the dataset provided by NIST and MITRE. The challenge was treated as a time series task in this experiment. It was observed that temporal Conv1D network performed better as compared to deep learning networks like Long Short-Term Memory (LSTM), ConvGRU and state of the art machine learning models like Support Vector Machines (SVM) and Decision tree-based models like XGBoost and Random Forest for their research study. The authors also highlighted several challenges while performing the experiment such as noise in the data distribution and poor transferability of the training data over the validation data which affects the accuracy of the model.\cite{b18}. 

\section{METHODOLOGY}

\begin{figure}[htbp] 
  \centering
  \includegraphics[scale = 0.3]{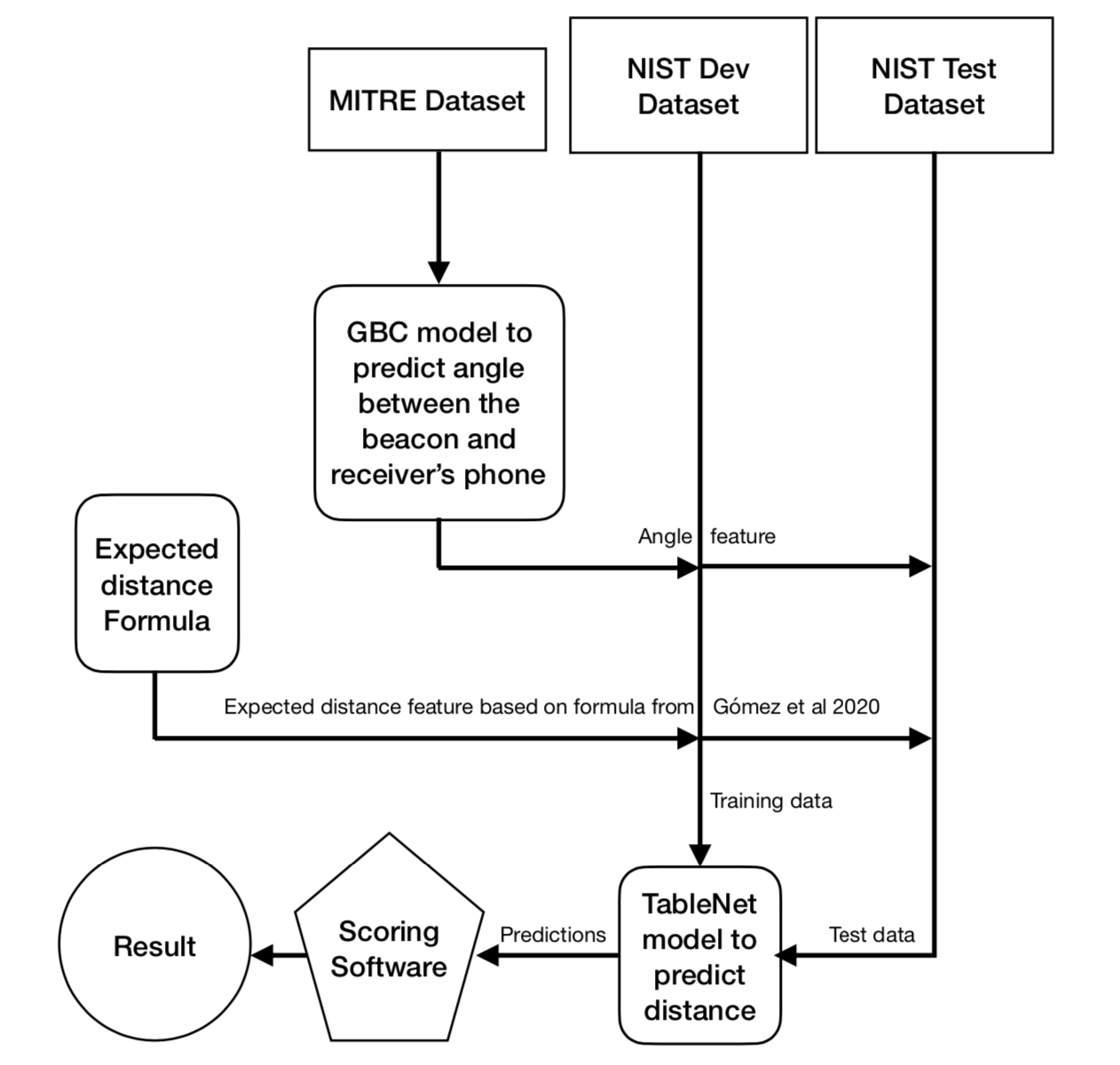}
  \caption{Experimental Workflow.}
  \small
  A high level overview of the experimental workflow.
  \label{perimental_flow}
\end{figure}

\subsection{Pipeline}
We made this problem as a four way classification task given an input features of mobile sensor data with classes being 1.2m, 1.8m, 3.0m and 4.5m. We choose classification over regression to predict distance because in both MITRE and TC4TL dataset distance label takes only those 4 values thus making it a appropriate choice. Our approach uses a two-stage pipeline, stage 1 and stage 2. Stage 1 prepares our engineered features using part of the input features. Stage 2 then uses those engineered features along with the residual input features to get results. An overview of the methodology with which we approached the challenge is illustrated in Fig. 2. 

\subsection{TableNet}
TableNet \cite{b21} is a deep learning model for end-to-end table detection and tabular data extraction from scanned documents. It uses VGG19 as its base network and then two decoder branches which uses features extracted from VGG19 to get desired segmented output for table region and column region. TableNet architecture have proven to be very effective in extracting information from scanned table documents. We modified TableNet so it can take feature vector as input and perform 4 way classification task.

\subsubsection{Stage 1} 
Engineered features are Expected Distance and Angle features. Expected Distance is calculated using Bluetooth signals received whereas the Angle feature is calculated using Attitude and Magnetic-Field values. Given the Bluetooth signals we calculate Expected Distance using equation \ref{eq2}.
Here the values of TX and N are taken as $-52$ and $2.6$ if the feature is coarse grain otherwise it is taken as $-54$ and $2.1$. These values are taken from \cite{b17}. For calculating Angle, we train a Gradient Boosting Classifier (GBC). Input features used were normalized Attitude and Magnetic-Field from MITRE dataset. Both the MITRE and NIST datasets were collected using the same protocol. Fig. 3. illustrates workflow of the pipeline used for training. 

\begin{equation}
d'=10^{\frac{TX-RSSI}{10*N}}\label{eq2}
\end{equation}

\begin{figure}[htbp]
\centerline{\includegraphics[scale=0.25]{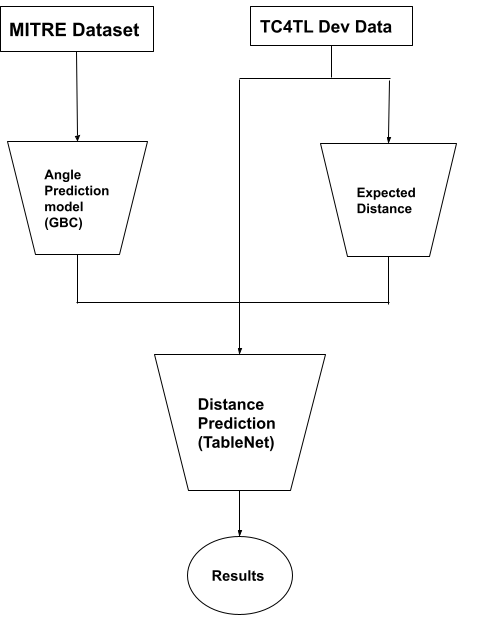}}
\caption{Pipeline used for training}
\label{training_pipeline}
\end{figure}

\subsubsection{Stage 2} 
\begin{figure}[htbp]
\centerline{\includegraphics[scale=0.25]{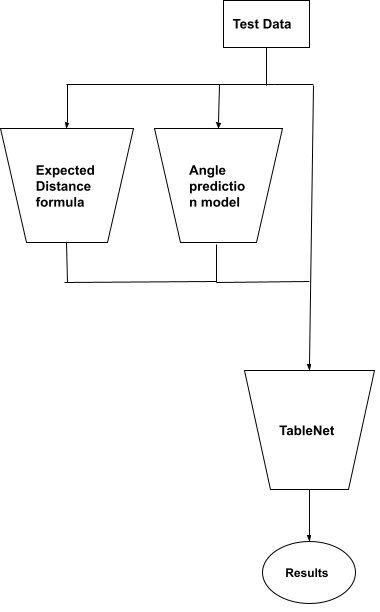}}
\caption{Pipeline used for test}
\label{test_pipeline}
\end{figure}

Stage 2 takes the input features along with the engineered features (i.e., Expected Distance and Angle) and predicts the distance. Input features used for Stage 2 are IMU sensor data (Gyroscope, Magnetic-field, Accelerometer, Attitude), phone specific bluetooth values (Bluetooth Power), beacon/transmitter metadata (Phone Carry Location, Pose), prediction requirement metadata (Coarse Grain Indicator) and engineered features (Expected Distance, Attenuation, Angle). For each event file we use pandas forward fill function to prepare our data using features mentioned above and remove rows with missing features. We also use attenuation as one of the features which is calculated using TXPower and Bluetooth values. Forward fill will propagate last valid observation forward to fill the missing values, that way each row has all the features necessary to predict the output. We use TableNet architecture for stage 2 to classify the distance for each row in an event file. Finally, we take mode of distances predicted over each row of the event file and assign that as resultant distance predicted by TableNet.\\
\textbf{Training} \\
We train GBC in stage 1 using no. of estimators as $100$ and learning rate of $0.2$ whereas TableNet was trained using Adam optimizer with learning rate of $0.02$ with learning scheduler using gamma value as $0.9$ and step size $10$. We trained TableNet for $1000$ epoch with batch size as $256$. Figure \ref{training_pipeline} shows our model pipeline for training whereas figure \ref{test_pipeline} is our inference pipeline. \\
\textbf{Testing} \\
For test we first evaluate angle feature using the given input feature. Later we use that feature along with the given input feature to get the distance class using TableNet. The scoring software computes $P\_miss$ and $P\_fa$ as well as $nDCF$ for distance values $1.2m$, $1.8m$ and $3.0m$ for both coarse grain and fine grain data. Fig. 4. illustrates workflow of the testing pipeline. 

\section{RESULTS AND DISCUSSIONS}

We propose a new state-of-the-art model, TDP (TableNet Distance Prediction). The model makes multiple predictions per event file using the most recent (past) sensor values along with the new measurements for each timestamp. Using the scoring software provided by TC4TL, we obtain the results shown in Table\ref{tab1}.Compared to previous TC4TL submissions, TDP outperforms previous best results as shown in Table \ref{tab2}. 

\begin{table}[]
\caption{Final results of our TDP model}
\begin{center}
\begin{tabular}{|c|c|c|c|c|}
\hline
\textbf{Subset}&\textbf{Distance}&\textbf{P\_miss}&\textbf{P\_fa}&\textbf{nDCF}  \\
\hline
\textbf{fine\_grain}& 1.20 & 0.01 & 0.02 & 0.02 \\
\hline
\textbf{fine\_grain}& 1.80 & 0.03 & 0.01 & 0.03 \\
\hline
\textbf{fine\_grain}& 3.00 & 0.01 & 0.14 & 0.15 \\
\hline
\textbf{coarse\_grain}& 1.80 & 0.01 & 0.01 & 0.01 \\
\hline
\textbf{average scores} & & 0.04 & 0.05 & 0.05 \\
\hline
\end{tabular}
\label{tab1}
\end{center}
\end{table}

\begin{table}[htbp]
\caption{nDCF scores for TDP \textit{VS} Previous TC4TL submissions}
\begin{center}
\begin{tabular}{|c|c|c|c|c|c|}
\hline
&\textbf{\shortstack{Fine \\ Grain \\ 1.2m}} & \textbf{\shortstack{Fine \\ Grain \\ 1.8m}} & \textbf{\shortstack{Fine \\ Grain \\ 3.0m}} & \textbf{\shortstack{Coarse \\ Grain \\ 1.8m}} & \textbf{\shortstack{Average \\ nDCF}}  \\
\hline
\textbf{GBM \cite{b17}}& 0.6 & 0.52 & 0.58 & 0.37 & 0.5175 \\
\hline
\textbf{\shortstack{Contact-Tracing \\ Project \cite{b17}}}& 0.68 & 0.54 & 0.59 & 0.41 & 0.555 \\
\hline
\textbf{TDP (ours)}& 0.02 & 0.03 & 0.15 & 0.01 & 0.05 \\
\hline
\end{tabular}
\label{tab2}
\end{center}
\end{table}

We further investigate the reason for our model performance by doing an ablation study. We check the dependency of our pipeline's results on each feature one by one by dropping them and noting the change in results on test set. 

\subsection{The effect of the Coarse Grain feature}\label{AA}
To examine the full effect of the Coarse Grain feature on the results, we removed it both as a feature in the model, and from use in the distance formula (as described above). We used the midpoint between the selected values for TX and N ($-53$, and $2.35$ respectively) to calculate the Expected Distance feature. The results of our model without this feature can be shown in Table \ref{tab3}. We found that the Coarse Grain feature has a strong positive effect on the model. This translated to the scoring software where we were getting a average $nDCF$ of $0.05$ with the Coarse Grain feature and $0.22$ without it. Another interesting feature of excluding the Coarse Grain feature is that the errors tend more towards missed positives, and the opposite is true when including it. Without the Coarse Grain feature the most important features in the model are Magnetic\_Field\_10, Magnetic\_Field\_12 and Attenuation. We feel this feature is worth including because of the positive effect it has on the model, and because, depending on how the algorithm is deployed, it will be known at the time of prediction if a coarse grained (effectively binary) result or a fine grained result is required. 

\begin{table}[htbp]
\caption{Results of TDP without Coarse Grain Feature}
\begin{center}
\begin{tabular}{|c|c|c|c|c|}
\hline
\textbf{Subset}&\textbf{D}&\textbf{P\_miss}&\textbf{P\_fa}&\textbf{nDCF}  \\
\hline
\textbf{fine\_grain}& 1.20 & 0.21 & 0.01 & 0.21 \\
\hline
\textbf{fine\_grain}& 1.80 & 0.14 & 0.07 & 0.21 \\
\hline
\textbf{fine\_grain}& 3.00 & 0.17 & 0.12 & 0.29 \\
\hline
\textbf{coarse\_grain}& 1.80 & 0.05 & 0.13 & 0.18 \\
\hline
\textbf{average scores} & & 0.14 & 0.08 & 0.22 \\
\hline
\end{tabular}
\label{tab3}
\end{center}
\end{table}

\subsection{The effect of the Expected Distance feature}
The Expected Distance feature had a small positive effect on the model. The results without the Expected Distance feature can be seen in Table \ref{tab4}. By dropping this feature we found our average $nDCF$ rose from $0.05$ to $0.06$. Without the Expected Distance feature the most important features in the model were Coarse\_Grain\_Y, Coarse\_Grain\_N and Attenuation. The model trained without the Expected Distance feature is much more likely to miss a positive case than predict a false positive compared to the model trained with the feature.

\begin{table}[htbp]
\caption{Results of TDP without Expected Distance Feature}
\begin{center}
\begin{tabular}{|c|c|c|c|c|}
\hline
\textbf{Subset}&\textbf{D}&\textbf{P\_miss}&\textbf{P\_fa}&\textbf{nDCF}  \\
\hline
\textbf{fine\_grain}& 1.20 & 0.00 & 0.02 & 0.02 \\
\hline
\textbf{fine\_grain}& 1.80 & 0.03 & 0.02 & 0.06 \\
\hline
\textbf{fine\_grain}& 3.00 & 0.02 & 0.11 & 0.13 \\
\hline
\textbf{coarse\_grain}& 1.80 & 0.01 & 0.00 & 0.01 \\
\hline
\textbf{average scores} & & 0.01 & 0.04 & 0.06 \\
\hline
\end{tabular}
\label{tab4}
\end{center}
\end{table}
\subsection{The effect of the Angle feature}
During model exploration and comparison, the Angle feature had a significant impact on performance. Our final model architecture, TableNet, however had enough training parameters to learn the angle feature from the raw sensor data. Including the Angle feature had a very small negative effect on the final model. The results without the Angle feature can be seen in Table \ref{tab5}. By dropping this feature we found our average $nDCF$ changed from $0.05$ to $0.04$. These values however fluctuated due to randomness of validation set splits, but it was decided to keep the Angle feature in the final model because it proved valuable for most other models. Without the Angle feature the most important features in the model (after Coarse Gain) became Attitude and Magnetic Field which were the features the Angle model was trained on. The model trained without the Angle feature is slightly more likely to miss a positive case than predict a false positive compared to the model trained with the feature.
\begin{table}[htbp]
\caption{Results of TDP without Angle Feature}
\begin{center}
\begin{tabular}{|c|c|c|c|c|}
\hline
\textbf{Subset}&\textbf{D}&\textbf{P\_miss}&\textbf{P\_fa}&\textbf{nDCF}  \\
\hline
\textbf{fine\_grain}& 1.20 & 0.01 & 0.01 & 0.03 \\
\hline
\textbf{fine\_grain}& 1.80 & 0.03 & 0.02 & 0.06 \\
\hline
\textbf{fine\_grain}& 3.00 & 0.03 & 0.07 & 0.10 \\
\hline
\textbf{coarse\_grain}& 1.80 & 0.01 & 0.01 & 0.02 \\
\hline
\textbf{average scores} & & 0.02 & 0.03 & 0.04 \\
\hline
\end{tabular}
\label{tab5}
\end{center}
\end{table}

\subsection{The effect of predicting on a subset of the event file}
Based on our hypothesis that the event files mimic the monotonic nature in which the data was collected, we applied the same model to three sets namely; the first look (first four seconds of event file), last look (last four seconds) and full event file. Surprisingly, the first look outperformed last look using average $nDCF$ ($0.0625$ and $0.19$ respectively). The full event file however resulted in the best average $nDCF$ using the same model and features ($0.06$). It should be noted that first  and last look had roughly the same amount of data points, whereas the full event had significantly more. Because of the results on first look, last look and full event file we suspect that true distance per timestamp may not be monotonic as stated in dataset documentation or there may be another reason which is not apparent yet. 

\subsection{The generalisability of the model}
Finally, given we had trained on the development set for the purpose of this challenge (making a prediction for each line in each event file), we wanted to test how generalisable the model would be when applied to the training set, which is less similar in structure to both the development and test sets.
\begin{table}[htbp]
\caption{Results of TDP with Training Set}
\begin{center}
\begin{tabular}{|c|c|c|c|c|}
\hline
\textbf{Subset}&\textbf{D}&\textbf{P\_miss}&\textbf{P\_fa}&\textbf{nDCF}  \\
\hline
\textbf{fine\_grain}& 1.20 & 0.67 & 0.38 & 1.05 \\
\hline
\textbf{fine\_grain}& 1.80 & 0.48 & 0.63 & 1.12 \\
\hline
\textbf{fine\_grain}& 3.00 & 0.12 & 0.89 & 1.02 \\
\hline
\textbf{coarse\_grain}& 1.80 & 0.39 & 0.58 & 0.97 \\
\hline
\textbf{average scores} & & 0.42 & 0.62 & 1.04 \\
\hline
\end{tabular}
\label{tab6}
\end{center}
\end{table}

\section{SUMMARY}
We proposed TDP model as our approach to the TC4TL challenge organised by NIST. We used a two stage pipeline which take advantage of dataset provided by TC4TL as well as MITRE dataset. We were able to improve on the existing state of the art models used to estimate the distance between the two phones for the NIST TC4TL using combinations of features with a TableNet model, which gives low false alarm rates and miss rates (total $nDCF$ $0.21$). The significant distinction between the model presented for this research study and previous attempts is that we predicted that angle at which the receiver was standing when the bluetooth signal was detected between two smartphones and used a TableNet model. As a result, this may be considered an important feature to predict for machine learning models being deployed as contact tracing apps.

\section{FUTURE WORK}
In our work we only used one additional dataset and two additional features from this dataset. In future, it may be beneficial to explore other potential features which could enhance the performance of the machine learning models. The MITRE Range Angle dataset, for example, includes a feature range that specifies the distance in feet from which the measurements were taken. As a result, it may be possible to train a model to predict the range for each dataset and use the predicted distance as the expected distance.

Based on our understanding of how the data is collected, we believe that future work should also investigate the use of different "looks." It remains unclear whether each event file is data collected at the same distance, such as is the case in the MIRTE Range Angle, and whether each “look” is capturing different angles or also different distances.

The model for predicting distance for this research study was trained using only the development dataset and not the NIST training dataset, which could be a limitation of the current approach. This dataset is smaller than the training dataset and more closely resembles the test dataset, raising the possibility of overfitting. Ideally, we could collect more data between the two phones and see how the model performs with the real world data.

TableNet is another area that we should investigate further. TableNet's is ideal for image data and generating $2d$ segmentation maps. It's performance on non image mobile sensor data for distance prediction is very interesting and it could be explored further. This also necessitates the use of other architectures available in the deep learning world and the testing for their performance.

\section*{ACKNOWLEDGMENT}
This work was funded by Science Foundation Ireland through the SFI Centre for Research Training in Machine Learning(18/CRT/6183).

\section{Code and Data}
Our code and data can be found on this Google Drive link:

\href{https://drive.google.com/drive/folders/1I0k3wM-ntMpirgoLS33j7eXLMpRWkpT5?usp=sharing}{\textcolor{blue}{Link}}


\begin{thebibliography}{00}

\bibitem{b1} Lu, H., Stratton, C. W., \& Tang, Y. W. (2020). Outbreak of pneumonia of unknown etiology in Wuhan, China: The mystery and the miracle. Journal of medical virology, 92(4), 401.
\bibitem{b2} World Health Organization, 2. (2020). WHO Director-General’s remarks at the media briefing on 2019-nCoV on 11 February 2020.
\bibitem{b3} World Health Organization. (2020). Novel Coronavirus ( 2019-nCoV), situation report, 11.
\bibitem{b4} Bebot Launches Free Coronavirus Information Bot (2020): https://www.be-spoke.io/index.html
\bibitem{b5} Baraniuk, C. (2020). Covid-19 contact tracing: a briefing. BMJ, 369.
\bibitem{b6} World Health Organization, \& Centers for Disease Control and Prevention
\bibitem{b7} Radcliffe, K., \& Clarke, J. (1998). Contact tracing—where do we go from here?.
\bibitem{b8} European Centre for Disease Prevention and Control. (2020). Contact tracing for COVID-19: current evidence, options for scale-up and an assessment of resources needed.
\bibitem{b9} Exposure Notification Bluetooth Specification, Apple|Google, https://covid19-static.cdn-apple.com/applications/covid19/current/static/contact-tracing/pdf/ExposureNotification-BluetoothSpecificationv1.2.pdf
\bibitem{b10} The PACT Protocol Specification https://pact.mit.edu/wp-content/uploads/2020/04/The-PACT-protocol-specification-ver-0.1.pdf
\bibitem{b11}  Leith, D. J., \& Farrell, S. (2020). Measurement-based evaluation of Google/Apple Exposure Notification API for proximity detection in a light-rail tram. Plos one, 15(9), e0239943.
\bibitem{b12} Kita, N., Ito, T., Yokoyama, S., Tseng, M. C., Sagawa, Y., Ogasawara, M., \& Nakatsugawa, M. (2009, March). Experimental study of propagation characteristics for wireless communications in high-speed train cars. In 2009 3rd European Conference on Antennas and Propagation (pp. 897-901). IEEE.
\bibitem{b13} Leith, D. J., \& Farrell, S. (2020). Coronavirus contact tracing: Evaluating the potential of using bluetooth received signal strength for proximity detection. ACM SIGCOMM Computer Communication Review, 50(4), 66-74.
\bibitem{b14} Hatke, G. F., Montanari, M., Appadwedula, S., Wentz, M., Meklenburg, J., Ivers, L., ... \& Fiore, P. (2020). Using Bluetooth Low Energy (BLE) signal strength estimation to facilitate contact tracing for COVID-19. arXiv preprint arXiv:2006.15711.
\bibitem{b15} L. Ceci. (2021). Mobile internet usage worldwide - statistics \& facts. https://www.statista.com/topics/779/mobile-internet/\#dossierKeyfigures
\bibitem{b16} Abbar, S., \& Mokbel, M. (2021). The role of AI in digital contact tracing. In Leveraging Artificial Intelligence in Global Epidemics (pp. 203-221). Academic Press.
\bibitem{b17} Gómez, C., Belton, N., Quach, B., Nicholls, J. and Anand, D. (2020). A Simplistic Machine Learning Approach to Contact Tracing. arXiv preprint arXiv:2012.05940.
\bibitem{b18} Shankar, S., Kanaparti, R., Chopra, A., Sukumaran, R., Patwa, P., Kang, M., ... \& Raskar, R. (2020). Proximity Sensing: Modeling and Understanding Noisy RSSI-BLE Signals and Other Mobile Sensor Data for Digital Contact Tracing. arXiv preprint arXiv:2009.04991.
\bibitem{b19}NIST TC4TL Challenge  https://www.nist.gov/itl/iad/mig/nist-tc4tl-challenge
\bibitem{b20}HE, T., and PRINTZ, M. (2020). A 2-stage Classifier for Contact Detection with BluetoothLE And INS Signals. NIST TC4TL Challenge.
\bibitem{b21} Paliwal, S. S., Vishwanath, D., Rahul, R., Sharma, M., \& Vig, L. (2019, September). Tablenet: Deep learning model for end-to-end table detection and tabular data extraction from scanned document images. In 2019 International Conference on Document Analysis and Recognition (ICDAR) (pp. 128-133). IEEE.
\bibitem{b22} NIST Pilot Too Close for Too Long (TC4TL) Challenge Evaluation Plan \url{https://www.nist.gov/system/files/documents/2020/07/01/2020_NIST_Pilot_TC4TL_Challenge_Evaluation_Plan_v1p3.pdf}

\end{thebibliography}
\end{document}